\documentclass[conference]{IEEEtran}
\ifCLASSINFOpdf
\else
\fi
\usepackage{times}
\usepackage{epsfig}
\usepackage{graphicx}
\usepackage{amsmath}
\usepackage{amssymb}
\begin{document}
%
\title{Dual Transformer for Point Cloud Analysis}

\author{\IEEEauthorblockN{Xian-Feng Han}
\IEEEauthorblockA{College of Computer and Information Science\\
Southwest University\\
}
\and

\IEEEauthorblockN{Yi-Fei Jin}
\IEEEauthorblockA{College of Computer and Information Science\\
Southwest University\\
}
\and
\IEEEauthorblockN{Hui-Xian Cheng}
\IEEEauthorblockA{College of Computer and Information Science\\
Southwest University\\
}

\and
\IEEEauthorblockN{Guo-Qiang Xiao}
\IEEEauthorblockA{College of Computer and Information Science\\
Southwest University\\
}
}


%


\maketitle

\begin{abstract}
Following the tremendous success of transformer in natural language processing and image understanding tasks, in this paper, we present a novel point cloud representation learning architecture, named Dual Transformer Network (DTNet), which mainly consists of Dual Point Cloud Transformer (DPCT) module. Specifically, by aggregating the well-designed point-wise and channel-wise multi-head self-attention models simultaneously, DPCT module can capture much richer contextual dependencies semantically from the perspective of position and channel. With the DPCT module as a fundamental component, we construct the DTNet for performing point cloud analysis in an end-to-end manner.  Extensive quantitative and qualitative experiments on publicly available benchmarks demonstrate the effectiveness of our proposed transformer framework for the tasks of 3D point cloud classification and segmentation, achieving highly competitive performance in comparison with the state-of-the-art approaches. 
\end{abstract}


%
\IEEEpeerreviewmaketitle

\section{Introduction}
\begin{figure*}[h]
\centering
   \includegraphics[width=\textwidth]{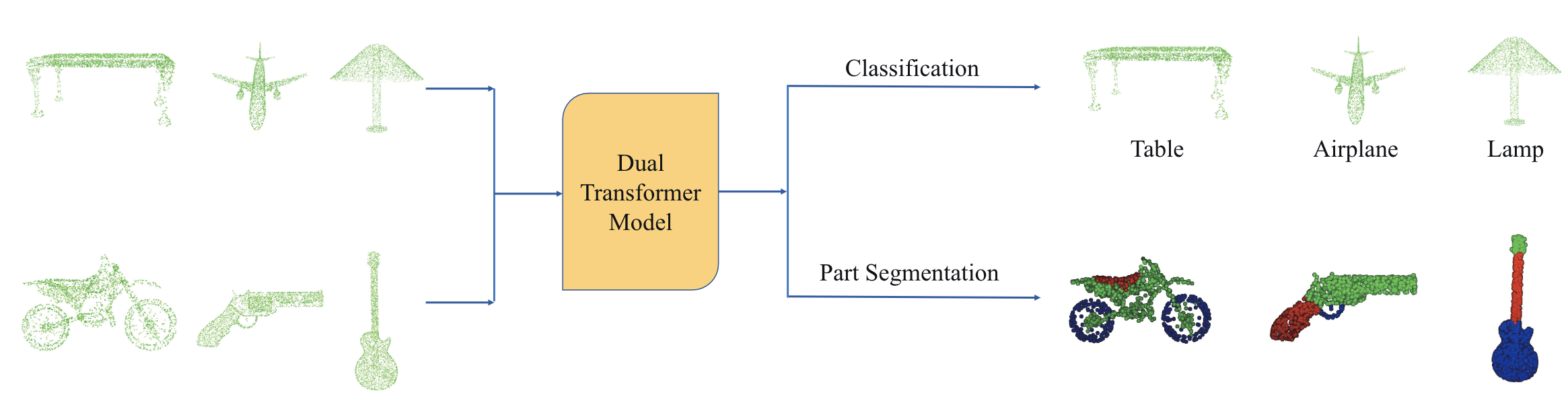}
  \caption{Applications of our proposed Dual Transformer architecture, including 3D object classification, part segmentation and semantic segmentation.}
  \label{fig:teaser}
\end{figure*}

The rapid development of 3D acquisition devices has witnessed the considerably important role of 3D point clouds played in a wide range of applications, such as virtual/augmented reality \cite{gojcic2020learning}\cite{jiang2020end}, autonomous driving \cite{guo2020deep}\cite{hu2020randla} and robotics \cite{nezhadarya2020adaptive}\cite{wang2019deep}. Therefore, how to perform effective and efficient point cloud analysis becomes an urgent requirement or an essential issue. Recently, deep learning techniques achieve tremendous success in 2D computer vision domain, which actually provide an opportunity for better understanding of point cloud. However, the inherent unordered, irregular and sparse structure makes it unreasonable to directly apply the traditional convolutional neural networks to 3D point clouds.

To address this problem, several existing state-of-the-art approaches attempt to transform the unstructured point cloud into either voxel grid \cite{maturana2015voxnet} or multi-views \cite{su2015multi}, then applying 3D CNN or 2D CNN for effective feature learning. Although these methods have reported promising performance, they suffer from growing memory requirement, high computational cost as well as geometric information loss during transformation. Alternatively, following the great success of pioneering work PointNet \cite{qi2017pointnet}, the family of Point-based approaches run directly on raw point cloud to learn 3D representation via share Multi-Layer Perceptrons (MLPs) \cite{qi2017pointnet++} or carefully-designed 3D convolutional kernel \cite{thomas2019kpconv}. Nevertheless, they may be insufficient to model long-range contextural correlation among points.

Recently, Transformer have achieved a series of breakthroughs in the field of Natural Language Processing (NLP) and 2D computer vision tasks \cite{chen2020generative}\cite{carion2020end}. As its core component, self-attention mechanism, on the one hand, is able to learn much richer contextural representation by capturing highly expressive long-range dependencies in the input. On the other hand, it is invariant to points permutations. Transformer, therefore, is an ideally mechanism suitable for point cloud processing.

Inspired by the superior performance of Transformer, we introduce a well-designed end-to-end architecture, named Dual Transformer Network (DTNet) for point cloud analysis. Its core component is our proposed Dual Point Cloud Transformer (DPCT) structure with the ability to aggregate the long-range spatial and channel context-dependencies of point-wise feature maps for enhancement of feature representation. Specifically, a DPCT module consists of two parallel branches. One is point-wise multi-head self-attention attempting to spatially extract contextural information among point-wise features, while the other is channel-wise multi-head self-attention model for capturing context dependencies in channel dimension. The output of these two attention modules are then concatenated via element-wise sum operation to improve the power of representation. Finally, we construct a U-Net like architecture using multi-scale DPCT modules to perform 3D point cloud analysis and understanding tasks.

Quantitative and quality evaluations conducted on challenging benchmark datasets demonstrate the effectiveness of our proposed model. And the superior performance further reflects the high competitiveness of our DTNet against other state-of-the-art point cloud learning networks on tasks of 3D shape classification, and segmentation.  

In summary, we make the following contributions:
\begin{itemize}
   
   \item A well-designed Dual Point Cloud Transformer (DPCT) module based on multi-heads self-attention is proposed for point cloud processing. This module can semantically enhance the representation power of learned point features by capturing long-range contextual dependencies from the position and channel correlations points of view.
   
   \item Based on our DPCT moduel, we construct an end-to-end Dual Transformer Network (DTNet) directly operating on 3D point cloud for highly effective feature learning.
   
   \item Extensive evaluations performed on three challenging benchmarks demonstrate the effectiveness of our point cloud Transformer architecture, achieving competitive performance on 3D object classification, and segmentation tasks.
\end{itemize}

\section{Related Work}

\subsection{Deep Learning on Point Cloud}
Motivated by the outstanding performance of deep learning techniques in 2D image analysis and understanding tasks, more and more attention has been attracted to extending deep learning on 3D point clouds for effective and efficient representations. However, it is unreasonable to directly apply the standard operations in CNNs to point clouds, largely due to its irregular, unordered and sparse structure. To address this issue, recent researches based on different data format have been proposed, which can be divided into volumetric-based, multi-view based, and point-based methods. 

\noindent\textbf{Volumetric-based methods}  \cite{maturana2015voxnet}\cite{le2018pointgrid} voxelize the unstructured point clouds into regular volumetric occupancy grid by quantization, which is well suitable for extracting feature representation with 3D convolutional neural networks. However, these approaches suffer from  cubically growing computational complexity and memory requirment with respect to volumetric resolution, as well as geometric information loss. To overcome these limitations, hierarchical data structure-based methods \cite{guo2020deep}, such as OctNet \cite{riegler2017octnet} and Kd-Net \cite{klokov2017escape}, have been proposed to concentrate on informative voxels while skipping the empty ones \cite{wang2019graph}. PointGrid \cite{le2018pointgrid} improves the local geometric details learning by incorporating points within each grid.

\noindent\textbf{Multi-view based methods} aims to turn 3D problem into 2D problem by projecting the point cloud space into a collection of 2D images with multiple views, so that 2D CNNs can be applied to perform feature learning. The pioneering work MVCNN\cite{MVCNN} simply leverages the max-pooling operation to extract multi-view features into a global descriptor. Wei et al.\cite{wei2020view} designed a directed graph by treating the views as graph nodes to construct the View-GCN. Although this type of methods have obtained remarkable performance on tasks like object classification \cite{qi2016volumetric}\cite{feng2018gvcnn}\cite{ma2018learning} with the well-established image CNNs, it is still difficult to determine the appropriate number of views to cover the 3D objects while avoiding information loss and high computational consumption.

\noindent\textbf{Point-based methods} directly take the raw point clouds as input for
learning 3D representations without any voxelization or projection. As a pioneering work, PointNet \cite{qi2017pointnet} learns point-wise feature directly from the raw point cloud via shared MLP and obtains global representation with max-pooling, which can achieve permutation invariance. Pointnet++\cite{qi2017pointnet++} extends PointNet by integrating a hierarchical structure that takes both global information and local details into consideration. Subsequently, the  works begin to focus on defining explicit convolution kernels for points. KPConv \cite{thomas2019kpconv} presents a deformable  point convolution using a collection of learnable kernel points. FPConv \cite{lin2020fpconv} runs directly on local surface of point cloud in projection-interpolation manner for efficient feature learning. Similar to these methods, our DTNet also deal with 3D point cloud directly without any intermediate grid or image representation.

\subsection{Transformers in vision}
More recently, Transformer networks\cite{jaderberg2015spatial} have made significant process in Natural Language Processing domain \cite{parmar2018image}\cite{hu2019local} and have achieved astounding performance. Its great success attracts increasing interest in computer vision researchers to transfer these models for 2D image tasks, such as classification \cite{dosovitskiy2020image} and detection \cite{carion2020end}. Actually, the core factor of Transformer is self-attention mechanism that has capability of explicitly modeling interactions between elements in the input sequence. iGPT \cite{chen2020generative} uses transformer for image generation tasks with unsupervised training mechanism. Vision Transformer (ViT) \cite{dosovitskiy2020image} replaces the standard convolution with transformers to perform large-scale supervised image classification. DETR \cite{carion2020end} is the first to attack the detection issue using transformer model from the perspective of prediction problem.  

Additionally, since self-attention does not depend on input order, and 3D point clouds can be treated as a set of points with positional attributes, transformer is ideally-suited for point cloud analysis. Therefore, we develop a point cloud transformer network with our carefully-designed Dual Point Cloud Transformer layer to correlate the point representations for learning broad geometry and context dependencies.

\section{Proposed Approach}
In this section, we propose a generic operation Dual Point Cloud Transformer (DPCT) that leverages the fundamental idea of transformer to model point sets interactions in terms of spatial and channel dependencies. By stacking the DPCT layers with increasing receptive field, we construct an end-to-end architecture called Dual Transform Network (DTNet), for point cloud analysis. The overall framework of our DTNet is shown in Figure \ref{fig_framework}.  

\subsection{Dual Point Cloud Transformer}
Given a D-dimension point cloud with $N$ points $\mathcal{P} = \{p_{i} \in \mathbf{R}^{D}, i = 1, 2, ..., N\}$, we aim to find a set function $f$ with property of permutation invariance to input and high expressiveness, which can project the input $\mathcal{P}$ to a long-range context enhancement feature space $\mathcal{F}$, $f: \mathcal{P} \longrightarrow \mathcal{F}$. Actually, Transformer and its associated self-attention mechanism are particularly suitable for this problem due to the usage of matrix multiplication, summation operations and ability of highlighting long-term relationships between elements. We, therefore, develop a Dual Point Cloud Transformer based on the advantage of transformer, which is composed of two individual multi-head self-attention branches for learning interactions across points and channels simultaneously.

\begin{figure*}[!t]
  \centering
  \includegraphics[width=7in]{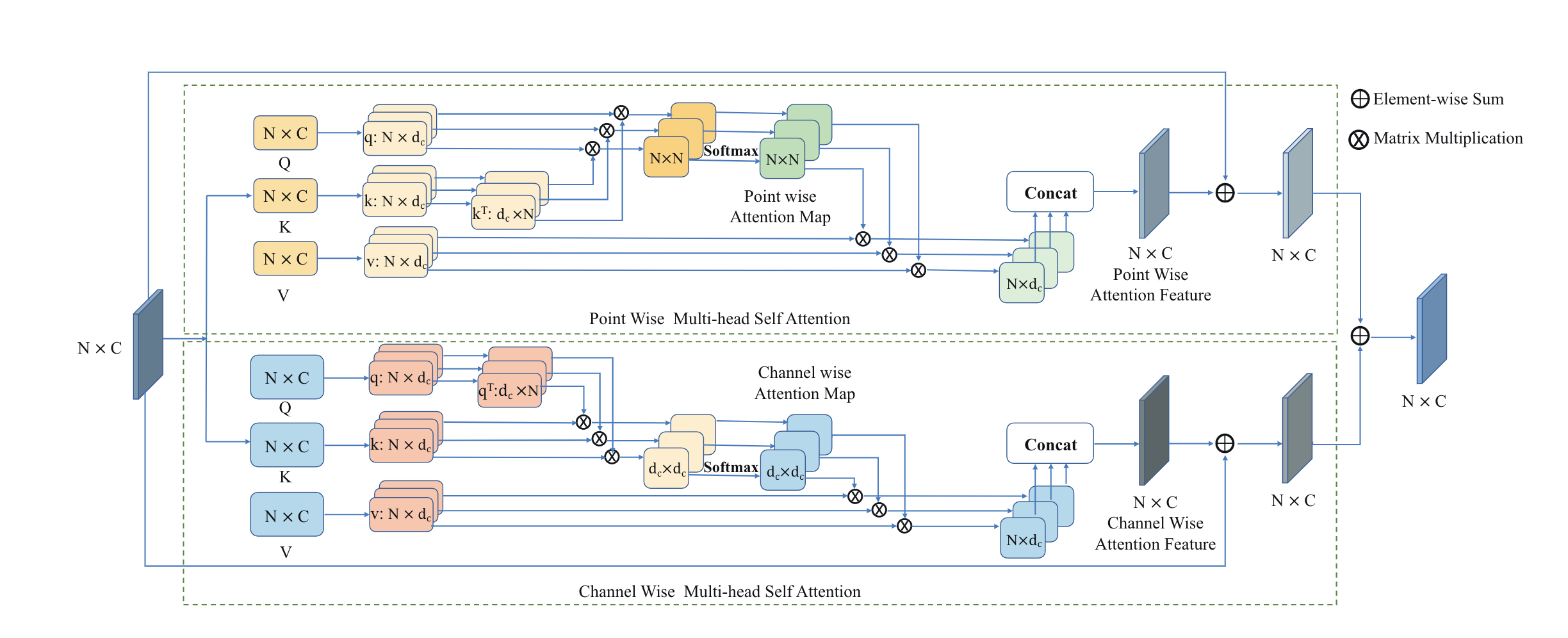}
  \caption{The details of our Dual Point Cloud Transformer.}
  \label{fig_dpct}
\end{figure*}

\noindent\textbf{Point-wise Self-attention.}In order to investigate the spatial correlation among points and generate long-range context-dependent representation, we construct a point-wise multi-head self-attention module for enhancement of feature capability.

As illustrated in the top of Figure \ref{fig_dpct}, consider a point-wise feature map from $l$th layer $F^{l} \in R^{N \times C}$, we formulated our attention model as:

\begin{equation}
    \mathcal{F}_{PWSA}^{l+1} = MHAT_{PWSA}(\mathcal{F}^{l}) = Concat(\mathcal{AT}^{1}_{l+1}, \mathcal{AT}^{2}_{l+1}, ..., \mathcal{AT}^{M}_{l+1}) + \mathcal{F}^{l}
\end{equation}
Where M denotes the number of self-attention blocks. $\mathcal{AT}(\cdot)$ operation is defined as:

\begin{equation}
    \mathcal{AT}^{m}(\mathcal{F}^{l}) = S^{m}_{l+1}V^{m}_{l+1} = \sigma(Q^{m}_{l+1}(K^{m}_{l+1})^{T} / \sqrt{C/M})V^{m}_{l+1}  
\label{equ:pwa}
\end{equation}

\begin{equation}
Q^{m}_{l+1} = \mathcal{F}^{l}W_{Q_{l+1}}^{m}
\end{equation}
\begin{equation}
 K^{m}_{l+1} = \mathcal{F}^{l}W_{K_{l+1}}^{m}
\end{equation}
\begin{equation}
 V^{m}_{l+1} = \mathcal{F}^{l}W_{V_{l+1}}^{m}
\end{equation}

The $m$ is the index of attention heads, $S^{m}_{l+1} \in R^{N \times N}$ is the point-wise attention matrix of the $m$th head, measuring the points' impacts on each other. $\sigma$ is the softmax operation. $W_{Q_{l+1}}^{m} \in R^{C \times d_{q}}$, $W_{K_{l+1}}^{m}\in R^{C \times d_{k}}$, $W_{V_{l+1}}^{m} \in R^{C \times d_{v}}$ are learnable weight parameters of three linear layers, where we set $d_{q} = d_{k} = d_{v} = d_{c} = C/M$.

From Equation \ref{equ:pwa}, we can conclude that the output point-wise feature map can be considered as a sum of features assembled from all points and the initial input, which spatially describes the long-range dependencies by integrating contextual information according to attention map.

\noindent\textbf{Channel-Wise Self-attention.} In order to highlight the role of interaction across different point-wise feature channels in improvement of contextural representation, we build the channel-wise multi-head self-attention model by adopting the similar strategy for calculating point-wise attention. As shown in the bottom of Figure \ref{fig_dpct}, the channel-wise attention is formally defined as:
\begin{equation}
        \mathcal{F}_{CWSA}^{l+1} = MHAT_{CWSA}(\mathcal{F}^{l}) = Concat(\mathcal{AT}^{'1}_{l+1}, \mathcal{AT}^{'2}_{l+1}, ..., \mathcal{AT}^{'M}_{l+1}) + \mathcal{F}^{l}
\end{equation}
\begin{equation}
    \mathcal{AT}^{'m}(\mathcal{F}^{l}) = U^{m}_{l+1}v^{m}_{l+1} = Softmax((q^{m}_{l+1})^{T}k^{m}_{l+1} / \sqrt{C/M})v^{m}_{l+1}  
\label{equ:cwa}
\end{equation}
\begin{equation}
q^{m}_{l+1} = \mathcal{F}^{l}W_{q_{l+1}}^{m}
\end{equation}
\begin{equation}
 k^{m}_{l+1} = \mathcal{F}^{l}W_{k_{l+1}}^{m}
\end{equation}
\begin{equation}
 v^{m}_{l+1} = \mathcal{F}^{l}W_{v_{l+1}}^{m}
\end{equation}
Where $U^{m}_{l+1} \in R^{d_{c} \times d_{c}}$ denotes the channel attention matrix, indicating the channel's importance to each other. $W_{q_{l+1}}^{m} \in R^{C \times d_{c}}, W_{k_{l+1}}^{m} \in R^{C \times d_{c}}, W_{v_{l+1}}^{m} \in R^{C \times d_{c}}$ are weight matrices of fully-connected layers. $d_{c} = C/M$.

Similar to point-wise attention, we sum the input and features across all channels to output the final channel-wise attention feature map. By leveraging the self-attention on channel level, we can encode much wider range of channel-wise relationships into representations to boost their capability.    

Finally, combining the above operations, our Dual Point Cloud Transformer generates a much more discriminative point-wise feature representation $\mathcal{F}^{l+1}_{DPCT}$containing long-range spatial and channel contextual information by performing element-wise addition on point-wise attention features and channel-wise attention features. 
\begin{equation}
    \mathcal{F}^{l+1}_{DPCT} = \mathcal{F}^{l+1}_{PWSA} \oplus \mathcal{F}^{l+1}_{CWSA}
\end{equation}

\subsection{Architecture of Dual Transformer Network}
\begin{figure*}[!t]
  \centering
  \includegraphics[width=7in]{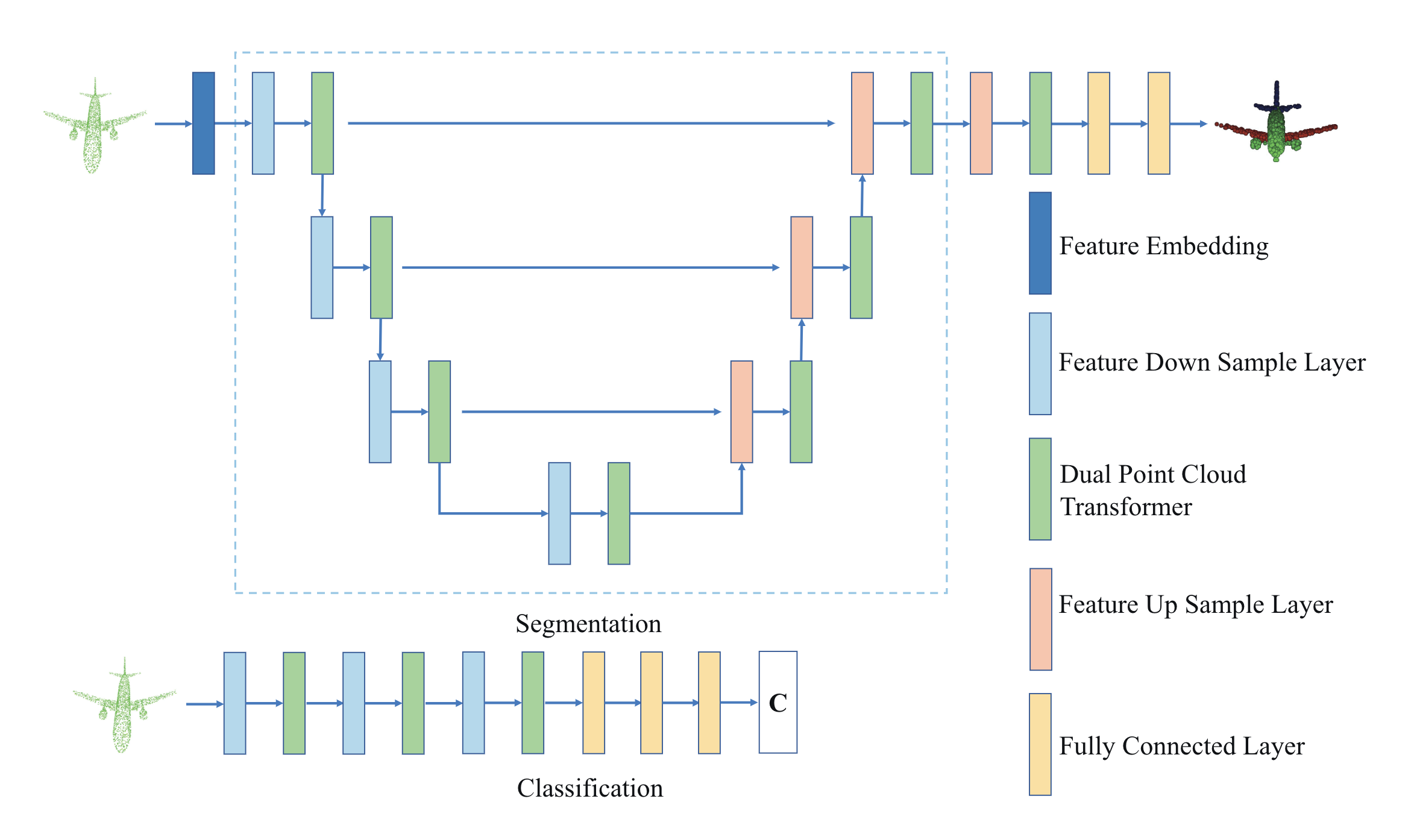}
  \caption{Dual Transformer Architecture for Point Cloud Analysis. The top network is for segmentation task, while the bottom is 3D object classification framework.}
  \label{fig_framework}
\end{figure*}

Based on the proposed Dual Point Cloud Transformer(DPCT) block, we construct complete deep networks for 3D point cloud analysis including classification and segmentation model, as shown in Figure \ref{fig_framework}. It can be obviously noted that the networks take point clouds as input, then progressively applying stacked Point Feature Network (PFN), DPCT layer, Feature Propagation (FP) Layer and Fully-connected(FC) layer, where our DPCT layer is the critical component for feature aggregation in our networks. 

\noindent\textbf{Feature Down Sample Layer.} In order to construct a hierarchical feature for multi-scale analysis, a Feature Down Sample (FDS) layer is added before our DPCT to downsample the point-wise feature maps as required.  Specifically, consider input $\mathcal{F}^{l}$, farthest point sampling algorithm (FPS) is performed to generate sub feature map $\mathcal{F}^{'l} \subset \mathcal{F}^{l}$, then we aggregate all point features in a r-ball neighboring region for each point in $\mathcal{F}^{'l}$, following by a linear transformation, batch normalization (BN) and ReLU operations. This FDS layer can be formally defined as:
\begin{equation}
    \mathcal{F}^{l+1} = Relu(BN(W^{'l}(Agg(FPS(\mathcal{F}^{l})))))  
\end{equation}
Where $Agg(\cdot)$ indicates the local feature aggregation operation. $W^{'l}$ denotes the learnable weight parameters of linear transformation.

\noindent\textbf{Feature Up Sample Layer.} In order to make dense prediction for segmentation tasks, we choose to put well-designed Feature Up Sample Layer in the decoder stage to progressively improve the resolution of point-wise feature map to the original size. K nearest neighbors interpolation is adopted to upsample point set according Euclidean distance.Similar to PointNet++ \cite{qi2017pointnet++}, we resort to skip connection and linear transformation together with batch normalization and ReLU for feature fusion between encoder and decoder. 

\noindent\textbf{Backbone.} Here, we make full use of the U-Net structure as basic backbone. Specifically, the number of layers in encoder and decoder stages, hyperparameter like downsampling rates are set particularly for each specific task and will be given detailedly in Experiment Section.

\section{Experiments}
In this section, we evaluate the effectiveness and performance of our DTNet on three publicly available datasets, namely ModelNet \cite{wu20153d}, and ShapNet \cite{yi2016scalable} , for tasks of classification, and segmentation, respectively. In addition, our point cloud transformer models are implemented with PyTorch framework on a single NVIDIA TITAN RTX 24G GPU. We use Adam strategy to optimize the networks. The weight decay is set to 0.0001.


\subsection{Point Cloud Classification}
\textbf{Dataset and metric.} We evaluate our 3D object classification transformer model on ModelNet40 \cite{wu20153d} which consists of 12,311 CAD models from 40 classes with 9,843 shapes for training and 2,468 objects for testing. Following Point Net \cite{qi2017pointnet}, 1,024 points are uniformly sampled from each model. During training, we perform data augmentation by adopting a random point dropout, a random scaling in [0.8, 1.25] and a random shift in [-0.1, 0.1]. Here, the overall accuracy (OA) is treated as evaluation metric for classification task.

\noindent\textbf{Network configuration.} Our Transformer network for shape classification task is shown in the bottom of Figure \ref{fig_framework}. We adopt three Point Feature Down Sample layers, each associating with an individual Dual Point Cloud Transformer module. Three fully-connected layers are appended at the end of our model. The number of points and channels used in each layer are summarized as follows: INPUT(N=1.024, C=3)-FDS(N=512, C=128)-DPCT(N=512, C=320)-FDS(N=128, C=256)-DPCT(N=256, C=640)-FDS(N=1, C=1024)-DPCT(C=1024)-FC(512)-FC(256)-FC(40).
The network is trained for 150 epcohs with batch size of 16 and an initial learning rate of 0.001 that is decayed by 0.7 every 20 epochs.

\noindent\textbf{Performance Comparison.} Performance of quantitative comparison with the state-of-the-arts are reported in Table \ref{tab:classification}. From the results, it can be clearly seen that our DTNet achieves the highest overall accuracy of 92.9\%, outperforming PointNet and the second best method Point2Sequence by 3.7\% and 0.3\%, respectively. This demonstrates the effectiveness of our model.

\begin{table}[htbp]
\centering
\caption{3D object classification results on ModelNet40. The best results are shown in bold.}
\begin{tabular}{cccc} \hline
Method & Representation   & Input Size    & ModelNet40 \\ \hline
3DShapeNets \cite{wu20153d} & Volumetric  & $30^{3}$ & 77.3\% \\
VoxNet \cite{maturana2015voxnet} &  Volumetric & $32^{3}$   & 83.0\% \\
MVCNN \cite{su2015multi}  & Multi-view & $12\times224^{2}$  & 90.1\%  \\
DeepNet \cite{ravanbakhsh2016deep} & Points & $5000\times3$  & 90.0\% \\
OctNet \cite{riegler2017octnet}  & Volumetric & $128^{3}$ &  86.5\% \\
Kd-Net \cite{klokov2017escape} & Points  & $2^{15} \times 3$ &  88.5\% \\
PointNet \cite{qi2017pointnet} & Points & $1024 \times 3$  & 89.2\%  \\
PointNet++ \cite{qi2017pointnet++}& Points+normals & $5000\times6$ & 91.9\%   \\
ECC \cite{simonovsky2017dynamic} & Points & 1000 $\times$3 &   83.2\% \\
DGCNN \cite{wang2018dynamic} & Points & $1024 \times 3$  & 92.2\% \\
PointCNN \cite{li2018pointcnn} & Points & $1024 \times 3$  & 92.5\% \\
KC-Net \cite{shen2018mining} & Points & $1024\times3$   & 91.0\% \\
FoldingNet \cite{yang2018foldingnet} & Points  & $2048\times3$ &  88.4\%  \\
Point2Sequence \cite{liu2019point2sequence} & Points & $1024\times3$ &  92.6\%   \\
OctreeGCNN \cite{lei2019octree} & Points &  $1024\times3$ & 92.0\%\\ 

SFCNN \cite{rao2019spherical} & Points+normals & $1024\times6$ &  92.3\% \\
3D-GCN \cite{lin2020convolution} & Points & $1024\times3$  & 92.1\%\\
ELM \cite{fujiwara2020neural} & Points & $1024\times3$ &  92.2\% \\
FPConv \cite{lin2020fpconv} & Points+normals & - &  92.5\% \\
SPH3D-GCN \cite{lei2020spherical} & Points & $1000 \times 3$ &  92.1\% \\
\hline
DTNet & Points & $1024 \times 3$ &  \textbf{92.9}\%\\
\hline
\end{tabular}

\label{tab:classification}
\end{table}

\noindent\textbf{Ablation Study and Analysis.} To further investigate the influence of our proposed transformer component, we carry out additional experiments on the classification task of ModelNet40. We remove the DPCT modules from the DTNet as the baseline. Table \ref{tab:ablation} presents the results of different design choices of DTNet in terms of accuracy average class (ACC) and overall accuracy (OA) results. 

\begin{table}
\centering
\caption{Ablation studies on the ModelNet40 dataset. We analyze the effects of point-wise self-attention (PWSA) and channel-wise self-attention (CWSA).}
\begin{tabular}{ccc} \hline
Method & ACC  & OA  \\ \hline
Baseline & 89.6 & 92.0 \\
Baseline + PWSA & 89.7 & 92.8 \\
Baseline + CWSA & 89.5 & 92.6 \\
DTNet &  \textbf{90.4}\% &  \textbf{92.9}\%\\
\hline
\end{tabular}

\label{tab:ablation}
\end{table}
Comparing with the baseline, we can clearly observe that point-wise and channel-wise self-attention make remarkable contributions to point cloud representation learning, achieving the results of 92.8\% and 92.6\% in OA, respectively. And our DPCT module with integration of these two attention models obtain a significant improvement over baseline by 0.9\%. The results convincingly validate the effectiveness and benefit of our DPCT block.

\subsection{Point Cloud Part Segmentation}
\textbf{Dataset and metric.} The object part segmentation task can be treated as a per-point classification problem. We choose to train and test our segmentation DTNet on ShapeNet Part benchmark dataset \cite{yi2016scalable}, which consists of 16,881 objects from 16 different classes with a total of 50 part-level labels. For experimental studies, we follow the officially defined 14,007/2,874 training/testing split. And 2,048 points are sampled from each shape as input. In addition, we perform the same data augmentation as that for classification. The standard evaluation metrics, including mean of IoU over all part classes and category-wise IoU, are computed to report the performance. 

\noindent\textbf{Network configuration.} The network architecture for part segmentation is presented in the top of Figure \ref{fig_framework}. In the encoder stage, the same structure as that for classification is used for multi-scale feature learning, while three feature propagation blocks (combination of Feature Up Sample layer and DPCT layer) are appended in decoder step. The detailed configurations for each layer are set as: INPUT(N=2048, C=3)-FDS(N=512, C=320)-DPCT(N=512, C=320)-FDS(N=128, C=512)-DPCT(N=128, C=512)-FDS(N=1, C=1024)-DPCT(N=1, C=2014)-FUS(N=128, C=256)-DPCT(N=128, C=256)-FUS(N=512, C=128)-DPCT(N=512, C=128)-FUS(N=2048, C=128)-DPCT(N=2048, C=128)-FC(128)-FC(50). We train our DTNet for 80 epochs. The initial learning rate is set to 0.0005 and decayed by half every 20 epoch. The batch size we use is 16.

\begin{figure*}[htbp]
  \centering
  \includegraphics[width=7in]{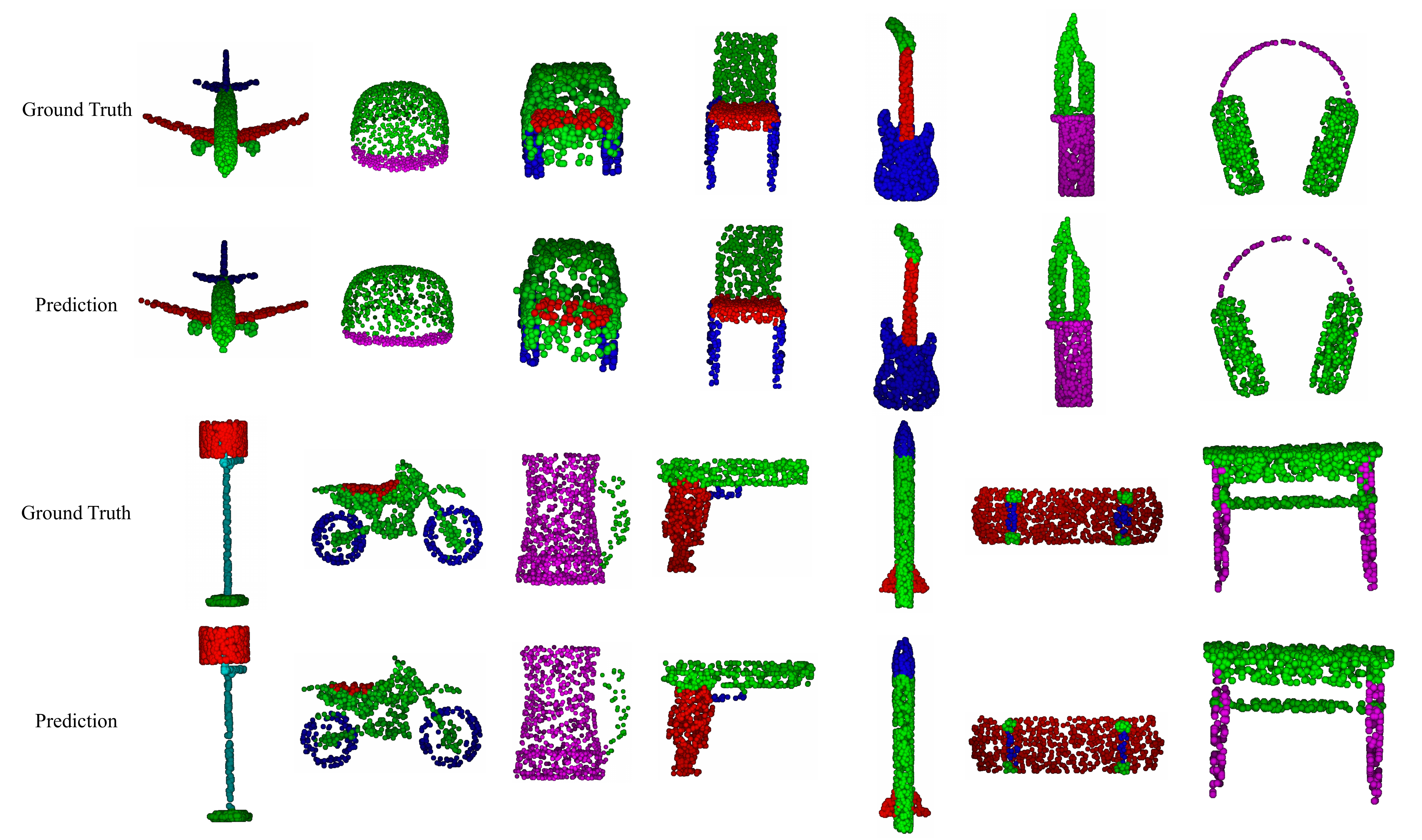}
  \caption{Qualitative comparison between our DTNet and the ground truth on ShapeNet Part dataset.}
  \label{fig_rsultforpart}
\end{figure*}

\noindent\textbf{Performance Comparison.} Quantitative comparison with previous state-of-the-art models are given in Table \ref{PartSegmentation}. Unlike PointNet, PointNet++ and SO-Net taking the surface normals together with point coordinates into account, our DTNet only uses XYZ coordinates as input feature. 
The segmentation result using our DTNet reaches the highest performance with respect to mIoU, achieving 85.6\% and surpassing PointNet++ and the previous best methods SFCNN, by 0.5\% and 0.2\%, respectively. In particularly, comparing with these competitive methods, our approach perform much better in some classes in terms of IoU, such as chair, lamp, skate, table etc. We visualize several 3D part segmentation in Figure \ref{fig_rsultforpart}. From these results, we can verify the success of our DTNet for part segmentation task.

\begin{table*}[h]
\caption{Part segmentation results on ShapeNet part dataset. The mean IoU across all the shape instances and IoU for each category are reported. }
\label{PartSegmentation}
\centering
\resizebox{\textwidth}{!}{
\begin{tabular}{c|c|cccccccccccccccc}
\hline
Method & mIoU & aero & bag & cap & car & chair & ep & guitar & knife & lamp & laptop & motor & mug & pistol & rocket & skate & table\\
\hline
ShapeNet \cite{yi2016scalable} & 81.4 & 81.0 & 78.4 & 77.7 & 75.7 & 87.6 & 61.9 & 92.0 & 85.4 & 82.5 & 95.7 & 70.6 & 91.9 & \textbf{85.9} & 53.1 & 69.8 & 75.3 \\
PointNet \cite{qi2017pointnet} & 83.7 & 83.4 & 78.7 & 82.5 & 74.9 & 89.6 & 73.0 & \textbf{91.5} & 85.9 & 80.8 & 95.3 & 65.2 & 93.0 & 81.2 & 57.9 & 72.8 & 80.6 \\
PointNet++ \cite{qi2017pointnet++} & 85.1 & 82.4 & 79.0 & 87.7 & 77.3 & 90.8 & 71.8 & 91.0 & 85.9 & 83.7 & 95.3 & 71.6 & 94.1 & 81.3 & 58.7 & 76.4 & 82.6\\
KD-Net \cite{klokov2017escape} & 82.3 & 80.1 & 74.6 & 74.3 & 70.3 & 88.6 & 73.5 & 90.2 & 87.2 & 71.0 & 94.9 & 57.4 & 86.7 & 78.1 & 51.8 & 69.9 & 80.3 \\
SO-Net \cite{li2018so} & 84.9 & 82.8 & 77.8 & 88.0 & 77.3 & 90.6 & 73.5 & 90.7 & 83.9 & 82.8 & 94.8 & 69.1 & 94.2 & 80.9 & 53.1 & 72.9 & 83.0 \\
RGCNN \cite{te2018rgcnn} & 84.3 & 80.2 & 82.8 & \textbf{92.6} & 75.3 & 89.2 & 73.7 & 91.3 & \textbf{88.4} & 83.3 & 96.0 & 63.9 & \textbf{95.7} & 60.9 & 44.6 & 72.9 & 80.4 \\
DGCNN \cite{wang2018dynamic} & 85.2 & \textbf{84.0} & 83.4 & 86.7 & 77.8 & 90.6 & 74.7 & 91.2 & 87.5 & 82.8 & 95.7 & 66.3 & 94.9 & 81.1 & \textbf{63.5} & 74.5 & 82.6\\
SRN \cite{duan2019structural} & 85.3 & 82.4 & 79.8 & 88.1 & 77.9 & 90.7 & 69.6 & 90.9 & 86.3 & 84.0 & 95.4 & \textbf{72.2} & 94.9 & 81.3 & 62.1 & 75.9 & 83.2 \\
SFCNN \cite{rao2019spherical} & 85.4 & 83.0 & 83.4 & 87.0 & \textbf{80.2} & 90.1 & 75.9 & 91.1 & 86.2 & 84.2 & \textbf{96.7} & 69.5 & 94.8 & 82.5 & 59.9 & 75.1 & 82.9 \\
3D-GCN \cite{lin2020convolution} & 85.1 & 83.1 & \textbf{84.0} & 86.6 & 77.5 & 90.3 & 74.1 & 90.9 & 86.4 & 83.8 & 95.6 & 66.8 & 94.8 & 81.3 & 59.6 & 75.7 & 82.6 \\ 
ELM \cite{fujiwara2020neural} & 85.2 & 84.0 & 80.4 & 88.0 & 80.2 & 90.7 & \textbf{77.5} & 91.2 & 86.4 & 82.6 & 95.5 & 70.0 & 93.9 & 84.1 & 55.6 & 75.6 & 82.1 \\ 
Weak Sup. \cite{xu2020weakly} & 85.0 & 83.1 & 82.6 & 80.8 & 77.7 & 90.4 & 77.3 & 90.9 & 87.6 & 82.9& 95.8 & 64.7 & 93.9 &79.8 & 61.9& 74.9 & 82.9 \\
\hline
DTNet & \textbf{85.6} & 83.0 & 81.4 & 84.3 & 78.4 & \textbf{90.9} & 74.3 & 91.0 & 87.3 & \textbf{84.7} & 95.6 & 69.0 &94.4 & 82.5 & 59.0 & \textbf{76.4} & \textbf{83.5}  \\

\hline
\end{tabular}}
\end{table*}

\section{Conclusion}
The recent advent of Transformer provides an solution to point permutation challenge faced by deep learning on point clouds. This paper introduced an end-to-end point cloud analysis transformer model, called Dual Transformer Network. Its core component is our well-designed Dual Point Cloud Transformer, which can capture long-range context dependencies by investigating the point-wise and channel-wise relationships. Extensive experiments on challenging benchmark datesets show the remarkable effectiveness of our DTNet, achieving state-of-the-art performance on tasks of classification, and segmentation. In the future, intensive study in Point Cloud Transformer should be made, including computational consumption reduction, design of new operations or networks and application to other tasks (semantic segmentation, shape completion, reconstruction,). 

\section*{Acknowledgment}



\bibliographystyle{IEEEtran}
\bibliography{IEEEabrv,paper}

\begin{thebibliography}{10}
\providecommand{\url}[1]{#1}
\csname url@samestyle\endcsname
\providecommand{\newblock}{\relax}
\providecommand{\bibinfo}[2]{#2}
\providecommand{\BIBentrySTDinterwordspacing}{\spaceskip=0pt\relax}
\providecommand{\BIBentryALTinterwordstretchfactor}{4}
\providecommand{\BIBentryALTinterwordspacing}{\spaceskip=\fontdimen2\font plus
\BIBentryALTinterwordstretchfactor\fontdimen3\font minus
  \fontdimen4\font\relax}
\providecommand{\BIBforeignlanguage}[2]{{%
\expandafter\ifx\csname l@#1\endcsname\relax
\typeout{** WARNING: IEEEtran.bst: No hyphenation pattern has been}%
\typeout{** loaded for the language `#1'. Using the pattern for}%
\typeout{** the default language instead.}%
\else
\language=\csname l@#1\endcsname
\fi
#2}}
\providecommand{\BIBdecl}{\relax}
\BIBdecl

\bibitem{gojcic2020learning}
Z.~Gojcic, C.~Zhou, J.~D. Wegner, L.~J. Guibas, and T.~Birdal, ``Learning
  multiview 3d point cloud registration,'' in \emph{Proceedings of the IEEE/CVF
  conference on computer vision and pattern recognition}, 2020, pp. 1759--1769.

\bibitem{jiang2020end}
H.~Jiang, F.~Yan, J.~Cai, J.~Zheng, and J.~Xiao, ``End-to-end 3d point cloud
  instance segmentation without detection,'' in \emph{Proceedings of the
  IEEE/CVF Conference on Computer Vision and Pattern Recognition}, 2020, pp.
  12\,796--12\,805.

\bibitem{guo2020deep}
Y.~Guo, H.~Wang, Q.~Hu, H.~Liu, L.~Liu, and M.~Bennamoun, ``Deep learning for
  3d point clouds: A survey,'' \emph{IEEE transactions on pattern analysis and
  machine intelligence}, 2020.

\bibitem{hu2020randla}
Q.~Hu, B.~Yang, L.~Xie, S.~Rosa, Y.~Guo, Z.~Wang, N.~Trigoni, and A.~Markham,
  ``Randla-net: Efficient semantic segmentation of large-scale point clouds,''
  in \emph{Proceedings of the IEEE/CVF Conference on Computer Vision and
  Pattern Recognition}, 2020, pp. 11\,108--11\,117.

\bibitem{nezhadarya2020adaptive}
E.~Nezhadarya, E.~Taghavi, R.~Razani, B.~Liu, and J.~Luo, ``Adaptive
  hierarchical down-sampling for point cloud classification,'' in
  \emph{Proceedings of the IEEE/CVF Conference on Computer Vision and Pattern
  Recognition}, 2020, pp. 12\,956--12\,964.

\bibitem{wang2019deep}
Y.~Wang and J.~M. Solomon, ``Deep closest point: Learning representations for
  point cloud registration,'' in \emph{Proceedings of the IEEE/CVF
  International Conference on Computer Vision}, 2019, pp. 3523--3532.

\bibitem{maturana2015voxnet}
D.~Maturana and S.~Scherer, ``Voxnet: A 3d convolutional neural network for
  real-time object recognition,'' in \emph{2015 IEEE/RSJ International
  Conference on Intelligent Robots and Systems (IROS)}.\hskip 1em plus 0.5em
  minus 0.4em\relax IEEE, 2015, pp. 922--928.

\bibitem{su2015multi}
H.~Su, S.~Maji, E.~Kalogerakis, and E.~Learned-Miller, ``Multi-view
  convolutional neural networks for 3d shape recognition,'' in
  \emph{Proceedings of the IEEE international conference on computer vision},
  2015, pp. 945--953.

\bibitem{qi2017pointnet}
C.~R. Qi, H.~Su, K.~Mo, and L.~J. Guibas, ``Pointnet: Deep learning on point
  sets for 3d classification and segmentation,'' in \emph{Proceedings of the
  IEEE conference on computer vision and pattern recognition}, 2017, pp.
  652--660.

\bibitem{qi2017pointnet++}
C.~R. Qi, L.~Yi, H.~Su, and L.~J. Guibas, ``Pointnet++: Deep hierarchical
  feature learning on point sets in a metric space,'' \emph{arXiv preprint
  arXiv:1706.02413}, 2017.

\bibitem{thomas2019kpconv}
H.~Thomas, C.~R. Qi, J.-E. Deschaud, B.~Marcotegui, F.~Goulette, and L.~J.
  Guibas, ``Kpconv: Flexible and deformable convolution for point clouds,'' in
  \emph{Proceedings of the IEEE/CVF International Conference on Computer
  Vision}, 2019, pp. 6411--6420.

\bibitem{chen2020generative}
M.~Chen, A.~Radford, R.~Child, J.~Wu, H.~Jun, D.~Luan, and I.~Sutskever,
  ``Generative pretraining from pixels,'' in \emph{International Conference on
  Machine Learning}.\hskip 1em plus 0.5em minus 0.4em\relax PMLR, 2020, pp.
  1691--1703.

\bibitem{carion2020end}
N.~Carion, F.~Massa, G.~Synnaeve, N.~Usunier, A.~Kirillov, and S.~Zagoruyko,
  ``End-to-end object detection with transformers,'' in \emph{European
  Conference on Computer Vision}.\hskip 1em plus 0.5em minus 0.4em\relax
  Springer, 2020, pp. 213--229.

\bibitem{le2018pointgrid}
T.~Le and Y.~Duan, ``Pointgrid: A deep network for 3d shape understanding,'' in
  \emph{Proceedings of the IEEE conference on computer vision and pattern
  recognition}, 2018, pp. 9204--9214.

\bibitem{riegler2017octnet}
G.~Riegler, A.~Osman~Ulusoy, and A.~Geiger, ``Octnet: Learning deep 3d
  representations at high resolutions,'' in \emph{Proceedings of the IEEE
  conference on computer vision and pattern recognition}, 2017, pp. 3577--3586.

\bibitem{klokov2017escape}
R.~Klokov and V.~Lempitsky, ``Escape from cells: Deep kd-networks for the
  recognition of 3d point cloud models,'' in \emph{Proceedings of the IEEE
  International Conference on Computer Vision}, 2017, pp. 863--872.

\bibitem{wang2019graph}
L.~Wang, Y.~Huang, Y.~Hou, S.~Zhang, and J.~Shan, ``Graph attention convolution
  for point cloud semantic segmentation,'' in \emph{Proceedings of the IEEE/CVF
  Conference on Computer Vision and Pattern Recognition}, 2019, pp.
  10\,296--10\,305.

\bibitem{MVCNN}
W.~Yin and H.~Sch{\"u}tze, ``Multichannel variable-size convolution for
  sentence classification,'' \emph{arXiv preprint arXiv:1603.04513}, 2016.

\bibitem{wei2020view}
X.~Wei, R.~Yu, and J.~Sun, ``View-gcn: View-based graph convolutional network
  for 3d shape analysis,'' in \emph{Proceedings of the IEEE/CVF Conference on
  Computer Vision and Pattern Recognition}, 2020, pp. 1850--1859.

\bibitem{qi2016volumetric}
C.~R. Qi, H.~Su, M.~Nie{\ss}ner, A.~Dai, M.~Yan, and L.~J. Guibas, ``Volumetric
  and multi-view cnns for object classification on 3d data,'' in
  \emph{Proceedings of the IEEE conference on computer vision and pattern
  recognition}, 2016, pp. 5648--5656.

\bibitem{feng2018gvcnn}
Y.~Feng, Z.~Zhang, X.~Zhao, R.~Ji, and Y.~Gao, ``Gvcnn: Group-view
  convolutional neural networks for 3d shape recognition,'' in
  \emph{Proceedings of the IEEE Conference on Computer Vision and Pattern
  Recognition}, 2018, pp. 264--272.

\bibitem{ma2018learning}
C.~Ma, Y.~Guo, J.~Yang, and W.~An, ``Learning multi-view representation with
  lstm for 3-d shape recognition and retrieval,'' \emph{IEEE Transactions on
  Multimedia}, vol.~21, no.~5, pp. 1169--1182, 2018.

\bibitem{lin2020fpconv}
Y.~Lin, Z.~Yan, H.~Huang, D.~Du, L.~Liu, S.~Cui, and X.~Han, ``Fpconv: Learning
  local flattening for point convolution,'' in \emph{Proceedings of the
  IEEE/CVF Conference on Computer Vision and Pattern Recognition}, 2020, pp.
  4293--4302.

\bibitem{jaderberg2015spatial}
M.~Jaderberg, K.~Simonyan, A.~Zisserman, and K.~Kavukcuoglu, ``Spatial
  transformer networks,'' \emph{arXiv preprint arXiv:1506.02025}, 2015.

\bibitem{parmar2018image}
N.~Parmar, A.~Vaswani, J.~Uszkoreit, L.~Kaiser, N.~Shazeer, A.~Ku, and D.~Tran,
  ``Image transformer,'' in \emph{International Conference on Machine
  Learning}.\hskip 1em plus 0.5em minus 0.4em\relax PMLR, 2018, pp. 4055--4064.

\bibitem{hu2019local}
H.~Hu, Z.~Zhang, Z.~Xie, and S.~Lin, ``Local relation networks for image
  recognition,'' in \emph{Proceedings of the IEEE/CVF International Conference
  on Computer Vision}, 2019, pp. 3464--3473.

\bibitem{dosovitskiy2020image}
A.~Dosovitskiy, L.~Beyer, A.~Kolesnikov, D.~Weissenborn, X.~Zhai,
  T.~Unterthiner, M.~Dehghani, M.~Minderer, G.~Heigold, S.~Gelly \emph{et~al.},
  ``An image is worth 16x16 words: Transformers for image recognition at
  scale,'' \emph{arXiv preprint arXiv:2010.11929}, 2020.

\bibitem{wu20153d}
Z.~Wu, S.~Song, A.~Khosla, F.~Yu, L.~Zhang, X.~Tang, and J.~Xiao, ``3d
  shapenets: A deep representation for volumetric shapes,'' in
  \emph{Proceedings of the IEEE conference on computer vision and pattern
  recognition}, 2015, pp. 1912--1920.

\bibitem{yi2016scalable}
L.~Yi, V.~G. Kim, D.~Ceylan, I.~Shen, M.~Yan, H.~Su, C.~Lu, Q.~Huang,
  A.~Sheffer, L.~Guibas \emph{et~al.}, ``A scalable active framework for region
  annotation in 3d shape collections,'' \emph{ACM Transactions on Graphics
  (TOG)}, vol.~35, no.~6, p. 210, 2016.

\bibitem{ravanbakhsh2016deep}
S.~Ravanbakhsh, J.~Schneider, and B.~Poczos, ``Deep learning with sets and
  point clouds,'' \emph{arXiv preprint arXiv:1611.04500}, 2016.

\bibitem{simonovsky2017dynamic}
M.~Simonovsky and N.~Komodakis, ``Dynamic edge-conditioned filters in
  convolutional neural networks on graphs,'' in \emph{Proceedings of the IEEE
  conference on computer vision and pattern recognition}, 2017, pp. 3693--3702.

\bibitem{wang2018dynamic}
Y.~Wang, Y.~Sun, Z.~Liu, S.~E. Sarma, M.~M. Bronstein, and J.~M. Solomon,
  ``Dynamic graph cnn for learning on point clouds,'' \emph{arXiv preprint
  arXiv:1801.07829}, 2018.

\bibitem{li2018pointcnn}
Y.~Li, R.~Bu, M.~Sun, W.~Wu, X.~Di, and B.~Chen, ``Pointcnn: Convolution on
  $\chi$-transformed points,'' in \emph{Proceedings of the 32nd International
  Conference on Neural Information Processing Systems}, 2018, pp. 828--838.

\bibitem{shen2018mining}
Y.~Shen, C.~Feng, Y.~Yang, and D.~Tian, ``Mining point cloud local structures
  by kernel correlation and graph pooling,'' in \emph{Proceedings of the IEEE
  conference on computer vision and pattern recognition}, 2018, pp. 4548--4557.

\bibitem{yang2018foldingnet}
Y.~Yang, C.~Feng, Y.~Shen, and D.~Tian, ``Foldingnet: Point cloud auto-encoder
  via deep grid deformation,'' in \emph{Proceedings of the IEEE Conference on
  Computer Vision and Pattern Recognition}, 2018, pp. 206--215.

\bibitem{liu2019point2sequence}
X.~Liu, Z.~Han, Y.-S. Liu, and M.~Zwicker, ``Point2sequence: Learning the shape
  representation of 3d point clouds with an attention-based sequence to
  sequence network,'' in \emph{Proceedings of the AAAI Conference on Artificial
  Intelligence}, vol.~33, no.~01, 2019, pp. 8778--8785.

\bibitem{lei2019octree}
H.~Lei, N.~Akhtar, and A.~Mian, ``Octree guided cnn with spherical kernels for
  3d point clouds,'' in \emph{Proceedings of the IEEE/CVF Conference on
  Computer Vision and Pattern Recognition}, 2019, pp. 9631--9640.

\bibitem{rao2019spherical}
Y.~Rao, J.~Lu, and J.~Zhou, ``Spherical fractal convolutional neural networks
  for point cloud recognition,'' in \emph{Proceedings of the IEEE/CVF
  Conference on Computer Vision and Pattern Recognition}, 2019, pp. 452--460.

\bibitem{lin2020convolution}
Z.-H. Lin, S.-Y. Huang, and Y.-C.~F. Wang, ``Convolution in the cloud: Learning
  deformable kernels in 3d graph convolution networks for point cloud
  analysis,'' in \emph{Proceedings of the IEEE/CVF Conference on Computer
  Vision and Pattern Recognition}, 2020, pp. 1800--1809.

\bibitem{fujiwara2020neural}
K.~Fujiwara and T.~Hashimoto, ``Neural implicit embedding for point cloud
  analysis,'' in \emph{Proceedings of the IEEE/CVF Conference on Computer
  Vision and Pattern Recognition}, 2020, pp. 11\,734--11\,743.

\bibitem{lei2020spherical}
H.~Lei, N.~Akhtar, and A.~Mian, ``Spherical kernel for efficient graph
  convolution on 3d point clouds,'' \emph{IEEE transactions on pattern analysis
  and machine intelligence}, 2020.

\bibitem{li2018so}
J.~Li, B.~M. Chen, and G.~H. Lee, ``So-net: Self-organizing network for point
  cloud analysis,'' in \emph{Proceedings of the IEEE conference on computer
  vision and pattern recognition}, 2018, pp. 9397--9406.

\bibitem{te2018rgcnn}
G.~Te, W.~Hu, A.~Zheng, and Z.~Guo, ``Rgcnn: Regularized graph cnn for point
  cloud segmentation,'' in \emph{Proceedings of the 26th ACM international
  conference on Multimedia}, 2018, pp. 746--754.

\bibitem{duan2019structural}
Y.~Duan, Y.~Zheng, J.~Lu, J.~Zhou, and Q.~Tian, ``Structural relational
  reasoning of point clouds,'' in \emph{Proceedings of the IEEE/CVF Conference
  on Computer Vision and Pattern Recognition}, 2019, pp. 949--958.

\bibitem{xu2020weakly}
X.~Xu and G.~H. Lee, ``Weakly supervised semantic point cloud segmentation:
  Towards 10x fewer labels,'' in \emph{Proceedings of the IEEE/CVF Conference
  on Computer Vision and Pattern Recognition}, 2020, pp. 13\,706--13\,715.

\end{thebibliography}
%

\end{document}